\documentclass[conference]{IEEEtran}
\usepackage{times}

\usepackage{multicol}
\usepackage[bookmarks=true]{hyperref}

\usepackage{url}

\usepackage{amsmath,amsfonts}
\usepackage{array}
\usepackage[caption=false,font=normalsize,labelfont=sf,textfont=sf]{subfig}
\usepackage{textcomp}
\usepackage{stfloats}
\usepackage{tabularray}
\usepackage{tabu}
\usepackage{booktabs} 
\usepackage{multirow}
\usepackage{multicol}
\usepackage{adjustbox}

\usepackage{xcolor}
\usepackage{xspace}
\usepackage{ifthen}

\usepackage{graphicx}
\usepackage{tikz}
\usetikzlibrary{positioning}

\let\labelindent\relax 
\usepackage{enumitem} 

\usepackage{amsmath,amssymb,amsfonts}
\usepackage{dsfont}
\usepackage{color}
\usepackage{balance}
\usepackage{gensymb}
\usepackage{siunitx} 
\usepackage{algorithm}
\usepackage{algorithmicx}
\usepackage{algpseudocode}
\usepackage{alphalph}
\usepackage{comment}
\usepackage{soul}
\usepackage{alphalph,etoolbox}

\graphicspath{{figures/}}

\allowdisplaybreaks[4]

\definecolor{navy}{rgb}{0,0,0.5}
\definecolor{dgreen}{rgb}{0,.7,0.6}
\definecolor{dyellow}{rgb}{.7,.7,0}
\definecolor{dred}{rgb}{1,0,0}
\definecolor{dblue}{rgb}{0,0,0.7}
\definecolor{brightblue}{rgb}{0.,0.5,1} 
\definecolor{dorange}{rgb}{0.9,0.5,0.1}
\definecolor{dgray}{rgb}{0.5,0.5,0.5}
\newcommand{\NEW}[1]{{\color{black} #1}}

\usepackage{diagbox}

\newcommand{\algoname}{Mirage\xspace}

\patchcmd{\subequations}{\alph{equation}}{\alphalph{\value{equation}}}{}{}

\algrenewcommand\algorithmicrequire{\textbf{Input:}}
\algrenewcommand\algorithmicensure{\textbf{Output:}}

\usepackage[utf8]{inputenc} 
\usepackage[T1]{fontenc}    
\usepackage{url}            
\usepackage{booktabs}       
\usepackage{amsfonts}       
\usepackage{nicefrac}       
\usepackage{microtype}      
\usepackage{tabularx}
\usepackage{microtype}
\usepackage{graphicx}
\usepackage{subcaption}

\usepackage[utf8]{inputenc} 
\usepackage[T1]{fontenc}    
\usepackage{dsfont}
\usepackage{nicefrac}       
\usepackage{color}
\usepackage{mathtools}
\usepackage{amsmath,amssymb}
\usepackage{bm}
\usepackage{siunitx}
\usepackage{wrapfig}
\usepackage{lipsum}
\usepackage{makecell}
\usepackage{subcaption}
\usepackage[capitalise, nameinlink]{cleveref}

\usepackage{footnote}
\makesavenoteenv{tabular}
\makesavenoteenv{table}

\usepackage{amsmath}
\usepackage{amssymb}
\usepackage{mathtools}

\usepackage{pifont}
\usepackage{graphicx} 
\usepackage{subcaption}

\usepackage{hyperref}
\usepackage{booktabs}
\usepackage{xspace}
\makeatletter
    \let\NAT@parse\undefined
\makeatother
\usepackage[square,numbers,sort&compress]{natbib}

\usepackage{hyperref}
\hypersetup{
    colorlinks=true,
    linkcolor=blue,
    filecolor=magenta,      
    urlcolor=magenta,
    citecolor=orange,
}
\usepackage[capitalise, nameinlink]{cleveref}

\begin{document}

\twocolumn
\newpage

\title{Mirage: Cross-Embodiment Zero-Shot Policy Transfer with Cross-Painting}

\author{
  Lawrence Yunliang Chen$^{*1}$,
  Kush Hari$^{*1}$,
  Karthik Dharmarajan$^{*1}$, 
  Chenfeng Xu$^1$,
  Quan Vuong$^2$,
  Ken Goldberg$^1$ \\ 
$^1$ UC Berkeley \quad\quad 
$^2$ Google DeepMind
\\ \href{https://robot-mirage.github.io/}{https://robot-mirage.github.io}
}

\makeatletter
\let\@oldmaketitle\@maketitle
\renewcommand{\@maketitle}{\@oldmaketitle
  \begin{center}
  \includegraphics[width=\textwidth]{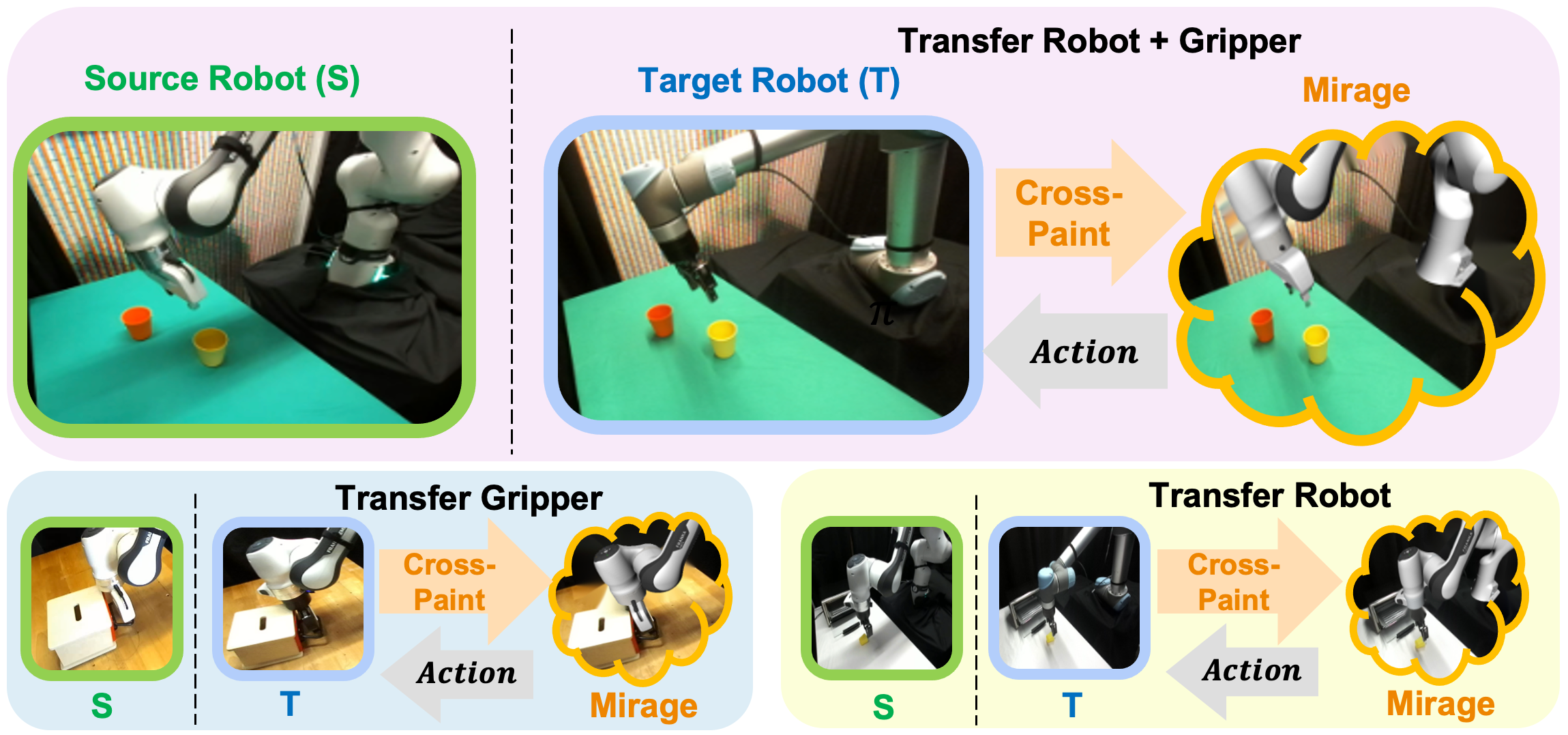}
    \captionof{figure}{Overview of \algoname. We study zero-shot policy transfer across embodiments. Assume there is a policy trained on a source robot (left). At test time, with an unseen target robot (middle), \algoname performs ``cross-painting''---masking out the target robot in the image and inpainting the source robot at the same end effector pose---using robot URDFs and a renderer. By creating an illusion as if the source robot were performing the task (right), \algoname queries the source policy with the cross-painted image to obtain the source robot's action. The target robot then uses a forward dynamics model to obtain the desired end effector pose in the target robot frame and executes the steps with a blocking controller. \algoname can successfully zero-shot transfer policies across the same robots with different grippers (bottom left), different robots with the same gripper (bottom right), and different robots with different grippers (top).} 
    \vspace{-5pt}
    \label{fig:teaser}
    \end{center}
}
\makeatother

\maketitle
\addtocounter{figure}{-1}
\begin{abstract}
The ability to reuse collected data and transfer trained policies between robots 
could alleviate the burden of additional data collection and training. While existing approaches such as pretraining plus finetuning and co-training show promise, they do not generalize to robots unseen in training. 
Focusing on common robot arms with similar workspaces and 2-jaw grippers, we investigate the feasibility of zero-shot transfer. 
Through simulation studies on 8 manipulation tasks, we find that state-based Cartesian control policies 
can successfully zero-shot transfer to a target robot after accounting for forward dynamics. 
To address robot visual disparities for vision-based policies,  
we introduce \algoname, which 
uses ``cross-painting''---masking out the unseen target robot and inpainting the seen source robot---during execution in real time so that it appears to the policy as if the trained source robot were performing the task. 
\NEW{\algoname applies to both first-person and third-person camera views and policies that take in both states and images as inputs or only images as inputs.}
Despite its simplicity, our extensive simulation and physical experiments provide strong evidence that \algoname can successfully zero-shot transfer between different robot arms and grippers with only minimal performance degradation on a variety of manipulation tasks such as picking, stacking, and assembly, significantly outperforming a generalist policy.

\end{abstract}
\IEEEpeerreviewmaketitle

\section{Introduction}
Consider a scenario where substantial efforts are invested in training a vision-based policy on a Franka robot arm for a manipulation task. The conventional paradigm of policy transfer to another robot often involves collecting new data on the target robot and finetuning the pretrained policy or retraining with a combination of datasets. Collecting data and training policies for each task and physical robot embodiment are time-consuming and expensive. 
An exciting conjecture is that demonstration data from different robot arms can be combined to learn policies that are robust or more generalizable to different robots, such as Franka, Universal, ABB, KUKA, and Fanuc arms~\cite{open_x_embodiment_rt_x_2023}. While scaling up shows great promise in in-distribution embodiments, transferring policies to unseen embodiments remains elusive.

In this work, we ask the question: Is it possible to achieve policy transfer without any target robot data? This poses several challenges, as outlined in prior work~\cite{yang2023polybot}, stemming from variations in kinematic configuration, control scheme, camera viewpoint, and end-effector morphology. However, commonalities exist among many robots in practice. Despite differences in joint numbers, many robots popular in open-sourced datasets~\cite{open_x_embodiment_rt_x_2023} such as the Franka, xArm, Sawyer, Kuka iiwa, and UR5 have similar workspaces and can be controlled in the Cartesian space of the end effector with millimeter-level accuracy. Similarly, while grippers may vary in appearance, most use parallel jaws with similar shapes (but differing dimensions). 

Our key insight is that, for robots with similar workspaces and many quasi-static tasks, the unseen target robot can achieve relatively high success rates by directly querying the policy trained on the seen source robot without the need for fine-tuning. To do so, we set the source robot to the same pose as the target robot using state information. The high success rates hold for policies that are both open-loop and closed-loop, both state-based and image-based, and trained using imitation learning and reinforcement learning.

We present \algoname (Fig.~\ref{fig:teaser}), a novel cross-embodiment policy transfer method that can zero-shot transfer policies trained on one source robot to an unseen target robot. \algoname decouples vision and control, allowing gaps between the source and target robot to be addressed separately. To address visual differences, \algoname employs ``cross-painting'' during execution, masking out the target robot and inpainting the source robot at the same pose, creating a ``mirage'' for the policy. 
Importantly, \algoname only requires knowledge of robot base coordinate frames for visual and state alignment. \algoname does not assume any demonstration data on the target robot or paired images or trajectories between the source and unseen target robots. \NEW{Moreover, \algoname applies to both first-person and third-person camera views, and policies that take in both states and images as inputs or only images as inputs.} To handle differences in control gains, \algoname pairs the source robot policy with a forward dynamics model and executes the action predicted by the policy on the target robot with a high-gain or blocking controller to accommodate varying control frequencies.

Through extensive experiments on 9 manipulation tasks in both simulation and real across 6 different robot and gripper setups, we show that \algoname, despite its simplicity, is remarkably effective at transferring policies, \NEW{achieving zero-shot performance significantly higher than a state-of-the-art generalist model~\cite{octo_2023}}. To the best of our knowledge, this is the first demonstration on real robots of zero-shot transfer of visual manipulation skills beyond pushing tasks~\cite{hu2021know}.

To summarize, our key contributions are:
\begin{enumerate}
    \item A systematic simulation study analyzing the challenges and potential for policy transfer between grippers and arms;
    \item \algoname, a novel zero-shot cross-embodiment policy transfer method that uses cross-painting to bridge the visual gap and forward dynamics to bridge the control gap;
    \item Physical experiments with Franka and UR5 demonstrating that \algoname successfully transfers between robots and grippers on 4 manipulation tasks, suffering only minimal performance degradation from the source policy and significantly outperforming a state-of-the-art generalist model. 
\end{enumerate}

\section{Related Work}
\subsection{Transfer across embodiments}
Can data and models from various sources, including other robots, tasks, environments~\cite{zhang2021policy, schubert2023generalist}, and modalities be transferred to new robots with differences in morphology, control, and sensors? While some cases pose significant disparities, such as a small soft robot with tactile sensors versus a humanoid robot with RGB cameras, commonly used robot arms share similarities in workspaces, grippers, and camera-based manipulation tasks.

\noindent \textbf{Transferring control dynamics.}
Previous work in control has investigated cross-robot transfer of dynamics models~\cite{chowdhary2013rapid, raimalwala2016preliminary, helwa2017multi, gao2021conformal}. Alignment-based transfer \cite{bocsi2013alignment, makondo2015knowledge, makondo2018accelerating} is a commonly used technique for trajectory tracking, where input-output data points for both robots are collected and then aligned using manifold alignment. The transformed data from the source robot can then be used as initialization to accelerate learning of the dynamics model of the target robot. \algoname does not focus on trajectory tracking. We fit a dynamics model to the source robot trajectory and use a high-gain or blocking controller on the target robot during execution.

\noindent \textbf{Transfer learning and cross-domain imitation} methods use finetuning to accelerate the learning process by leveraging data from other robots~\cite{taylor2009transfer, hu2019skill, zhu2023transfer, salhotra2023bridging}. For example, assuming access to a policy of the source robot, one can use Reinforcement Learning (RL) to finetune the visual encoder~\cite{rusu2017sim, sun2022transfer}, the value function~\cite{konidaris2006autonomous}, or the policy~\cite{liu2022revolver, liu2022meta} on the target robot. When learning from videos of other agents without access to actions~\cite{schmeckpeper2021reinforcement, yu2018one, bonardi2020learning, smith2019avid, liu2018imitation, xiong2021learning, xu2023xskill, schmeckpeper2021reinforcement, bahl2022whirl, wen2023any}, methods such as explicit retargeting~\cite{peng2020learning, sivakumar2022robotic}, inverse RL~\cite{zakka2022xirl, chen2021learning}, and goal-conditioned RL on a learned latent space~\cite{zhou2021manipulator} have been explored. Assuming isometry between domains, \citet{fickinger2021cross} train RL in the target domain by using a trajectory in the source domain as a pseudo-reward. Many other works assume data for both source and target robots on a proxy task, learn correspondences if not already available~\cite{ammar2015unsupervised, kim2020domain, zhang2020learning, raychaudhuri2021cross}, and then use the learned latent space or paired trajectories as auxiliary rewards~\cite{lakshmanan2010transfer, ammar2015unsupervised, gupta2017learning, shankar2022translating} or for adversarial training~\cite{hejna2020hierarchically, franzmeyer2022learn, yin2022cross}. Unlike these methods, \algoname does not assume access to target robot data on a proxy task or paired trajectories, and directly reuses the source robot policy.

\noindent \textbf{Multi-task and Multi-robot training.}
If the target robot is unknown but comes from a known distribution, such as the distribution of the length of the arm links, kinematic domain randomization~\cite{exarchos2021policy} can be used to train a robot-agnostic policy. Alternatively, the robot parameters such as kinematic parameters~\cite{yu2023multi}, 6 DOF transforms for each joint~\cite{chen2018hardware}, and URDFs represented by a stack of point clouds~\cite{shao2020unigrasp} or TSDF volumes~\cite{xu2021adagrasp} can be used to train a robot-conditioned policy. ~\citet{ghadirzadeh2021bayesian} use meta-learning to infer the latent vector for a new robot from few-shot trajectories, and ~\citet{noguchi2021tool} treat grasped tools as part of the embodiment to learn to grasp objects with objects. Other work has also leveraged a known robot distribution to train modular policies~\cite{devin2017learning, furuta2022system, Zhou2023Modularity, jian2023policy, xiong2023universal}. For example, ~\citet{devin2017learning} train a policy head for each task and each robot, and ~\citet{furuta2022system} train an RL policy for each task and each morphology combination and distill it into a transformer. Others~\cite{wang2018nervenet, sanchez2018graph, pathak2019learning, malik2019zero, huang2020one, kurin2020my} encode robot morphology as a graph for locomotion. While these methods leverage simulation to vary the robots in the known distribution, we 
do not seek to train a policy with the intention of it being performant on a distribution of robots. Instead, \algoname can transfer a specialized policy trained on only one robot to an unseen target robot that the policy is not intended to work on.

~\citet{yang2023polybot} study both the control and visual domain gaps when transferring across robots. They use wrist cameras to minimize the observation differences of the robot, learn a multiheaded policy with robot-specific heads that capture separate dynamics, and use contrastive learning to align robot trajectories. In contrast, \algoname works with \NEW{both first-person and third-person cameras} and does not assume any target robot data to jointly train a multi-robot policy.
Another closely related work to ours is Robot-Aware Visual Foresight~\cite{hu2021know}. The authors use the known camera matrix and robot model to mask the robot pixels, train a video-prediction model that only predicts the world pixels, and use visual foresight~\cite{finn2017deep, ebert2018visual} during execution. Similar to their method, we use robot URDFs to compute robot pixels, but we not only mask out the target robot but also inpaint the source robot. By doing so, we do not impose restrictions on the policy class the source robot can use, 
allowing \algoname to work with state-of-the-art source robot policies such as diffusion policy~\cite{chi2023diffusionpolicy} and successfully transfer them across real robots beyond pushing tasks~\cite{hu2021know}.

\subsection{Learning from large datasets}
Beyond targeted effort to transfer robot policies, recent work has also explored use of large and diverse data~\cite{depierre2018jacquard, kalashnikov2018qt, levine2018learning, acronym2020, shafiullah2023dobbe, fang2023rh20t, ebert2021bridge, walke2023bridgedata} to train visual encoders~\cite{nair2022r3m, xiao2022masked, ma2022vip} and policies that are generalizable to new objects, scenes, tasks, as well as embodiments~\cite{alayrac2022flamingo, jang2022bc, lynch2023interactive, shridhar2022cliport, stone2023moo, shridhar2022peract, jiang2022vima, reed2022a, radosavovic2022real, shah2023gnm, shah2023vint, roboagent, brohan2023rt1, brohan2023rt2, chen2023palix, driess2023palme}. 
~\citet{dasari2019robonet} use a large dataset of multiple robots to pretrain a visual dynamics model. Most recently, several works~\cite{radosavovic2023robot, bousmalis2023robocat, open_x_embodiment_rt_x_2023, octo_2023} pretrain large transformers and show that co-training or finetuning the models outperform policies trained only on in-domain data. However, RPT~\cite{radosavovic2023robot} uses joint space actions, preventing it from zero-shot transfer. While RT-X~\cite{open_x_embodiment_rt_x_2023} and Octo~\cite{octo_2023} use Cartesian space actions, they do not align action spaces or condition on the robot coordinate frames and kinematics, preventing them from working zero-shot on a new robot setup as they do not know what action space to use. In this work, we conduct experiments that show that leveraging the robot URDFs and aligning the action spaces could enable even a single-robot specialist policy to generalize zero-shot to a different robot. 

A work that is closely related to ours is MimicGen~\cite{mandlekar2023mimicgen}. Given novel object poses, they spatially transform human demonstration trajectories in an object-centric way and let the robot follow these generated trajectories to collect new demonstrations. They illustrate that MimicGen can be used to generate rollout data on Sawyer, IIWA, and UR5e arms even when the original demonstrations were on the Franka, and the generated data can be used to learn a target robot policy. In contrast, \algoname does not assume an object-centric framework and uses cross-painting for trained policies, eliminating the requirements for demonstration trajectories on the target robot.

\subsection{Image inpainting and augmentation in robotics}
Visual domain randomization such as adding distractors and changing the textures and lighting is commonly used to bridge the sim2real gap~\cite{tobin2017domain}. Recently, enabled by generative models, researchers have explored image editing such as adding distractors and changing backgrounds as an additional form of augmentation~\cite{mandi2022cacti, chen2023genaug, yu2023scaling}. 
\citet{black2023zero} use a video prediction model to generate goal images during execution. \citet{hirose2023exaug} augment the collected data to simulate another robot’s observation from a different viewpoint. AR2-D2~\cite{duan2023ar2} renders a virtual AR robot in a scene observed by a camera to replace the hand of a human manipulating objects; the result appears as if the AR robot were manipulating the object. \citet{bahl2022whirl} use a video inpainting model to mask out human hands and robot grippers. \algoname also inpaints the source robot, and can optionally reproject the image into the camera angle the policy is trained on, to create the illusion to the policy as if the source robot were performing the task.

\section{Problem Statement}

We assume two robot arms, source $\mathcal{S}$ and target $\mathcal{T}$, with parallel-jaw grippers, known URDFs, and known isomorphic kinematics. We assume the manipulation tasks of interest are in both robots' workspaces and can be performed by both robots using similar strategies without collision. 
Our goal is to transfer a trained policy from the source robot to the target robot.

Prior work~\cite{yang2023polybot} has found aligning the action and observation spaces can facilitate policy transfer.
\algoname leverages the following assumptions and design choices to reduce the gap between robots and enable zero-shot transfer:

\begin{enumerate}
    \item We assume knowledge of the two robots' coordinate frames. To align the state and action spaces, we follow prior work~\cite{yang2023polybot} and use the Cartesian space of the end effector. This allows us to transfer between robots with different numbers of joints and compensate for alternate gripper shapes across embodiments. 
    \item We assume the target robot has a high-gain or blocking controller that can reach desired poses relatively accurately (within a few millimeters).
    \item We assume knowledge of the camera parameters for both domains. This allows us to render robots in a camera pose that is within the distribution of the training image poses. 
    \item We assume that the background and lighting conditions of the target robot are in the distribution of the source dataset $D$ or that the policy $\pi_{\mathcal{S}}$ is robust to environmental background changes, e.g., using techniques such as background augmentation~\cite{mandi2022cacti, chen2023genaug, yu2023scaling}. This allows us to separate any challenges that arise due to changes in the background environment and focus on the impact of visual differences between robots on policy performance. 
\end{enumerate}

We model each robot manipulation task of interest as a Markov Decision Process.
We consider the setting where there is a policy $\pi_\mathcal{S}$ trained on a dataset of the source robot $D = \{(s_1^{\mathcal{S}}, o_1^{\mathcal{S}}, a_1^{\mathcal{S}}, ..., s_{H_i}^{\mathcal{S}}, o_{H_i}^{\mathcal{S}})_i^N\}$ consisting of $N$ trajectories, where $s_t^{\mathcal{S}}$ is the proprioceptive state of the robot at timestep $t$, $o_t^{\mathcal{S}}$ is the \NEW{first- or third-person camera observation(s)}, and $a_t^{\mathcal{S}}$ is the action. 
Given a source policy action $a_{t+1}^{\mathcal{S}} = \pi_\mathcal{S}(s_t^{\mathcal{S}}, o_t^{\mathcal{S}})$, we would like to transform it into a target policy action $a_{t+1}^{\mathcal{T}} = \pi_\mathcal{T}(s_t^{\mathcal{T}}, o_t^{\mathcal{T}})$ that takes as inputs the states and observations of the target robot without demonstration or finetuning trajectory data on the target robot.


\begin{figure*}
    \centering
    \includegraphics[width=0.95\linewidth]
    {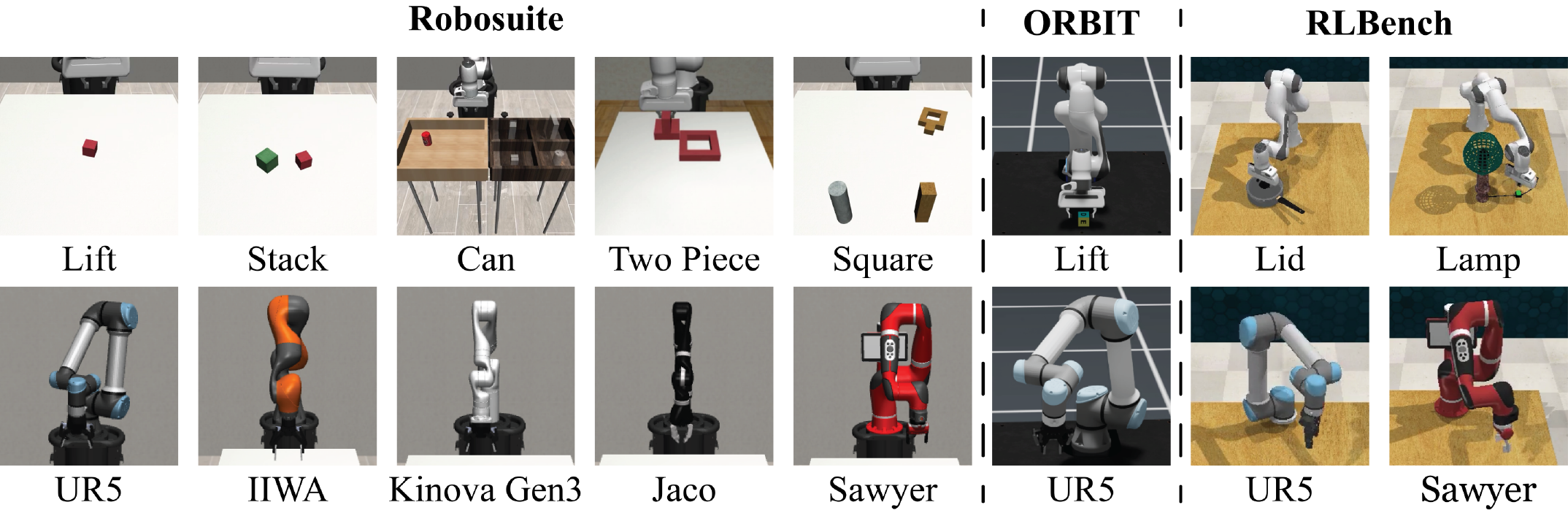}
    \caption{Simulation Tasks and Robots. The simulation evaluation utilizes the Robosuite simulator with Lift, Stack, Can Pick-and-Place, Two Piece Assembly, and Square Peg Insertion tasks. Additionally, to study other policy classes, we evaluate policy transfers in ORBIT with a block lifting task, and in RLBench with 2 tasks: Lifting a lid, and Pushing a button to turn on a lamp. For all policies, the source robot is the Franka robot as shown in the top row, while the target robots for each of the tasks are shown in the bottom row.}
    \label{fig:sim_tasks}
    \vspace{-5pt}
\end{figure*}

\section{State-Based Transfer Experiments}\label{sec:sim_study}
As a first step in studying the feasibility of zero-shot transfer, we seek to separate the domain gaps of the control from the visuals. 
To motivate the study, imagine there is a source robot (``oracle'') teaching a target robot to perform a task side by side in a duplicate environment. At each timestep, the source robot sees the world state $(p_{\mathrm{r}}, p_{\mathrm{o}})$ of the target environment, where $p_{\mathrm{r}}$ and $p_{\mathrm{o}}$ are the poses of the robot end effector and the objects. Then, it puts its objects and end effector to the same poses, and uses its policy to move its end effector to a new pose $p_{\mathrm{r}}'$. The target robot observes the source robot and also moves its end effector there. In this manner, the target robot mirrors what the source robot does step by step to perform the task.
We ask: can the target robot successfully complete a task by querying the source robot's policy in this fashion?

To answer this question, we consider 8 tasks across 3 simulators (Robosuite~\cite{robosuite2020}, ORBIT~\cite{mittal2023orbit}, and RLBench~\cite{james2020rlbench}) (Fig.~\ref{fig:sim_tasks}) with policies trained using imitation learning for Robosuite tasks and reinforcement learning for ORBIT and RLBench. Robosuite and ORBIT policies use closed-loop control (a trajectory consists of $>$50 timesteps), while for RLBench, the policies use open-loop control (a trajectory consists of a few waypoints). For all tasks, we train the source state-based policy on the Franka robot and evaluate the success rates on different target robots using the test-time execution strategy mentioned above. 

As the coordinate frames between robots are not necessarily the same, we use the known rigid transform $T_{\mathcal{T}}^{\mathcal{S}}$ between the frames to convert the end-effector and object poses at each time step from the target frame to the source frame $p_{\mathrm{r}}^\mathcal{S} = T_{\mathcal{T}}^{\mathcal{S}} p_{\mathrm{r}}^\mathcal{T}$ and $p_{\mathrm{o}}^\mathcal{S} = T_{\mathcal{T}}^{\mathcal{S}} p_{\mathrm{o}}^\mathcal{T}$, before querying the source robot's policy $a^{\mathcal{S}} = \pi_\mathcal{S}(p_{\mathrm{r}}^{\mathcal{S}}, p_{\mathrm{o}}^{\mathcal{S}})$. We step the action $a^{\mathcal{S}}$ on the source robot in the simulator to obtain the achieved end effector pose ${p_{\mathrm{r}}^{\mathcal{S}}}'$, and then use the high-gain or blocking controller on the target robot to reach the equivalent desired pose ${p_{\mathrm{r}}^\mathcal{T}}' = T_{\mathcal{S}}^{\mathcal{T}} {p_{\mathrm{r}}^\mathcal{S}}'$.


\begin{table*}[htbp!]
\centering
{\begin{tabular}{c|c|ccccc|ccccc}
\toprule[1pt]
\multirow{2}{*}{\textbf{Task}} & \multirow{2}{*}{\textbf{Source (Franka)}} & \multicolumn{5}{c}{\textbf{Franka Gripper on Target Robot}}  & \multicolumn{5}{c}{\textbf{Default Gripper on Target Robot}} \\

    \cmidrule(lr){3-7} \cmidrule(lr){8-12}

& & UR5e & IIWA & Kinova Gen3 & Sawyer & Jaco & UR5e & IIWA & Kinova Gen3 & Sawyer & Jaco$^*$ \\
\midrule[0.1pt]
\midrule[0.1pt]
Lift (Robosuite) & 100\% & 99\% & 100\% & 100\% & 100\% & 100\% & 100\% & 100\% & 99\% & 84\% & 95\% \\

Stack & 95\% & 96\% & 94\% & 96\% & 94\% & 96\% & 86\% & 87\% & 81\% & 85\% & 66\% \\

Can & 98\% & 98\% & 97\% & 97\% & 96\% & 96\% & 88\% & 90\% & 83\% & 83\% & 55\% \\

Two Piece Assembly & 96\% & 91\% & 89\% & 88\% & 92\% & 90\% & 89\% & 70\% & 92\% & 90\% & 34\% \\

Square & 81\% & 74\% & 72\% & 34\% & 83\% & 21\% & 69\% & 48\% & 49\% & 75\% & 1\% \\

\midrule[0.1pt]
Lift (ORBIT) & 100\% & - & - & - & - & - & 89\% & - & - & - & - \\

\midrule[0.1pt]
Unlid Pan & 100\% & - & - &  -& - & - & 100\% & - & - &  90\% & - \\

Lamp On & 88\% & - & - & - & - & - & 64\% & - & - & 68\% & - \\

\bottomrule[1pt]
\end{tabular}} 
\caption{\textbf{State-Based Policy Transfer Experiment Results.} We evaluate state-based policies trained for each task using a Franka robot across five different robots equipped either with the original Franka gripper or with each target robot's default gripper. Results suggest that most unseen target robots can successfully perform the tasks using the source robot as its guide for where to move its gripper. $^*$Jaco has a 3-jaw gripper, which explains its lower success rates.}
\label{tab:control-study-table}
\vspace{-5pt}
\end{table*}

\subsection{Implementation Details}
For Robosuite, we choose 5 tasks: Lift, Stack, Can, Two Piece Assembly, and Square. We use Robomimic~\cite{robomimic2021} to train an LSTM policy for each task on the provided demonstration data~\cite{robosuite2020, mandlekar2023mimicgen}. We evaluate the performances on 5 different robots, including UR5e, Kuka iiwa, Kinova Gen3, Sawyer, and Jaco. 
Note that Jaco has a 3-jaw gripper, but we include it for comparison. The source policy predicts delta Cartesian actions and we use the operation space controller~\cite{khatib1987unified} on the target robot to servo to the pose the source robot reaches and enforce that the norm of the error from the desired pose (position (in \si{\meter}) and quaternion) is less than 0.015 at each timestep.

For ORBIT, we use the Lift task. We train an RL policy using PPO~\cite{schulman2017proximal} on the Franka robot and evaluate on the UR5. Similar to Robosuite, we use the absolute pose controller to servo to the desired pose at each step during execution. For RLBench, we use Coarse-to-Fine Q-attention~\cite{james2022coarse} to learn the key poses and use its end-effector-pose-via-planning controller to reach the desired pose in an open-loop fashion. The policy also takes in camera observations and the scene point cloud. We study 2 tasks: take the lid off a saucepan (``Unlid Pan'') and turn on a lamp (``Lamp On''), and evaluate on the UR5 and Sawyer.

\subsection{Study Results}\label{subsec:sim_study_results}
Table~\ref{tab:control-study-table} shows that when the target robots have the same gripper as the source robot, most unseen target robots achieve very high task success rates. This suggests that the kinematic differences among the robot arms are relatively insignificant. Using the robots' default factory 2-jaw grippers, there is a mild performance drop of around 10-25\%, but many target robots can still successfully perform the task using the source robot as its guide for where to move its gripper. In comparison, with a 3-jaw gripper, Jaco's success rates are significantly lower than the others', especially on more challenging tasks, where the grasp configuration required for three jaws is different than the parallel jaw grippers. 
Comparing the robots, we see that the differences in the performance are roughly consistent across tasks, indicating that gripper properties (e.g., size and friction of the gripper pads) affect how easy it is for the robots to grasp and manipulate objects, but the general task strategy is similar. This holds for policies trained using IL and RL, as well as open loop and closed loop. 

Additionally, we notice that the more robust the source policy is, the smaller the performance drop when transferring to other robots. We qualitatively observe that this can be attributed to the extra space the policy leaves between the object and its gripper as well as its retrying behavior. Less robust source policies leave little room for error, while more robust ones tend to retry even if the target robot fails to grasp the object the first time.

To study the impact of differences in the robot controller dynamics, we also experiment with executing the delta action $a$ on the target robot with the non-blocking controller directly instead of using a blocking controller to reach the desired pose ${p_{\mathrm{r}}^\mathcal{T}}'$. Results are included in the Appendix. We see that there is a significant drop in performance, indicating that the difference in the forward dynamics between robots cannot be ignored when transferring policies, and that leveraging a blocking controller on the target robot is an effective way to mitigate this difference.

\section{\algoname: A Cross-Embodiment Transfer Strategy for Vision-Based Policies}

Motivated by the observation that target robots can successfully perform tasks to a large extent simply by querying a state-based source policy on an oracle source robot that mirrors its pose and obtains the next pose, we seek to extend this strategy to vision-based policies. To transfer vision-based policies, we need to account for the additional difference of robot visuals in addition to the controller forward dynamics. 

We propose \algoname, a strategy to zero-shot transfer a trained vision-based policy from the source robot to the target robot. The key idea is ``cross-painting'': replacing the target robot with the source robot in the camera observations at test time so that it appears to the policy as if the source robot were performing the task.

\subsection{Bridging the Visual Gap}
\begin{figure*}[t]
\center
\includegraphics[width=1.00\textwidth]{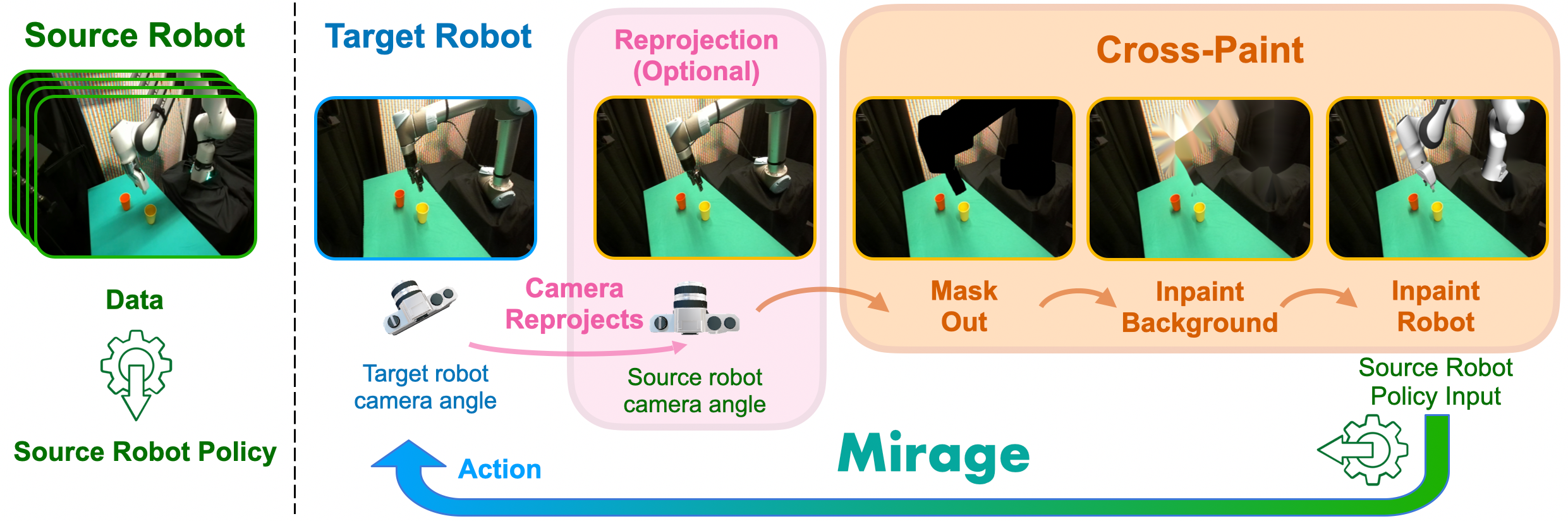}
\caption{
\NEW{\textbf{Illustration of \algoname's pipeline}. We reproject the camera from the target frame to the source frame if there is a non-negligible camera angle change and then apply cross-painting: (1) use the segmentation mask provided by Gazebo to mask out the target robot, (2) apply the fast marching \cite{telea2004image} algorithm to fill in the missing pixels, and (3) overlay Gazebo's rendering of the source robot URDF onto the image. The resulting image is fed into the source robot's policy to obtain the action, which is executed after a coordinate frame transform.}
}
\label{fig:inpainting}
\end{figure*}

To replace the robots, we leverage the knowledge of the robot URDFs and camera poses to perform cross-painting at test time. \NEW{Fig.~\ref{fig:inpainting} illustrates the pipeline of \algoname.} First, given known camera transforms, we can optionally reproject the images from the target domain to the source domain if depth sensing is available at test time. Next, given the image observation $o^\mathcal{T}$ and joint angles of the target robot, we use a renderer to determine which image pixels correspond to the source robot and mask out these pixels. Then, we inpaint the missing pixels. We use the fast marching inpainting method~\cite{telea2004image,opencv_library} for simplicity and speed, but other choices such as off-the-shelf image or video inpainting networks~\cite{yu2023inpaint, lee2019copy} could also work. Another potential approach is to first take an offline picture of the background scene with as much of the target robot arm moved out of the camera frame as possible, and at test time just to fill in the masked-out region with the corresponding pixels from the background image, but this would only apply to fixed third-person cameras. Finally, we use the URDF of the source robot to solve for the joint angles that would put its end effector at the same pose as that of the target robot, render it using a simulator, and overlay it onto the target image. For simulation experiments, we take into account potential occlusions between the robot and objects by comparing the pixel-wise depth values between the camera observation of the scene and the rendered robot. For real experiments, however, we do not use depth due to noise and imprecision in the camera observations. For the gripper, we similarly compute and set the joints of the source robot gripper in the renderer so that its width would roughly match that of the target robot's gripper. We denote this cross-painted image $o^{\mathcal{T}\rightarrow\mathcal{S}}$.

\subsection{Bridging the Control Gap}
To bridge the difference of the controllers, we use a forward dynamics model to convert the source robot action into the next pose to achieve, and use a high-gain or blocking controller on the target robot to reach the pose. We use the source robot trajectory data $D$ to fit a forward dynamics model $f$ on the transitions: $f(p_{\mathrm{r},t}^{\mathcal{S}}, a_t^{\mathcal{S}}) = p_{\mathrm{r},t+1}^{\mathcal{S}}$. During the target robot's execution, the desired target pose is thus $p_{\mathrm{r},t+1}^\mathcal{T} = T_{\mathcal{S}}^{\mathcal{T}} p_{\mathrm{r},t+1}^\mathcal{S} = T_{\mathcal{S}}^{\mathcal{T}} f(p_{\mathrm{r},t}^{\mathcal{S}}, a_t^{\mathcal{S}}) = T_{\mathcal{S}}^{\mathcal{T}} f(p_{\mathrm{r},t}^{\mathcal{S}}, \pi_\mathcal{S}(p_{\mathrm{r,t}}^{\mathcal{S}}, o^{\mathcal{T}\rightarrow\mathcal{S}}))$.

\section{Vision-Based Policy Transfer Experiments} \label{sec:mirage_exps}

We aim to answer the following questions:
\begin{enumerate}
    \item Can cross-painting bridge the visual gap between robots?
    \item Can \algoname successfully zero-shot transfer trained vision-based policies from one robot to another? \NEW{Does \algoname work with different camera views and policies?}
    \item To what extent does each component of \algoname affect the transfer performance? 
\end{enumerate}

\subsection{Simulation Experiments}\label{subsec:sim_experiments}

\begin{table*}
\centering
\begin{tabular}{c|c| ccc| ccc}
    \toprule
\multirow{2}{*}{\textbf{Task}}  & \multirow{2}{*}{\textbf{Source (Franka)}} & \multicolumn{3}{c}{\textbf{UR5e (Franka / Default Gripper)}}  & \multicolumn{3}{c}{\textbf{Kinova Gen3 (Franka / Default Gripper)}}  \\
    \cmidrule(lr){3-5} \cmidrule(lr){6-8}  
        & & Oracle  &  \NEW{\algoname\,-- CP} & \algoname  & Oracle  & \NEW{\algoname\,-- CP} & \algoname         \\
    \midrule
    \midrule
     Lift & 97\% & 98\% / 98\% & 55\% / 0\% & 88\% / 76\%  &  100\% / 97\% & 48\% / 1\% & 88\% / 72\% \\
    
    Stack & 97\% & 97\% / 95\% & 17\% / 5\%  & 84\% / 80\% & 99\% / 96\% & 12 \% / 2\% & 76\% / 80\% \\
    
    Can & 94\% & 71\% / 56\% & 13\% / 1\% & 68\% / 48\%  & 58\% / 44\% & 15\% / 2\% & 40\% / 32\% \\
    
    Two Piece Assembly & 99\% & 97\% / 99\% &  16\% / 2\% & 96\% / 100\% &  89\% / 91\% & 32\% / 25\% & 84\% / 80\% \\
    
    Square & 70\% & 73\% / 69\% & 22\% / 3\% & 68\% / 52\% & 48\% / 32\% & 8\% / 2\% &  40\% / 40\% \\
    \bottomrule 
\end{tabular}
\caption{\textbf{\algoname Results on Transferring State-And-Vision-Based Policies in Simulation.} For each task and robot arm combination, \NEW{the source BC-RNN policy uses both robot gripper poses and images as inputs.} Oracle represents the performance of the policy assuming access to a ground truth rendering of the source robot given the state of the target robot. \NEW{\algoname\,-- CP ablates cross-painting and directly passes the visual observation of the target robot to the policy.} \algoname uses cross-painting to generate the visual inputs for the policy. For each method, the first number represents the success rate when the target robot uses the source robot (Franka) gripper and the second number corresponds to using the target robot's default gripper (Robotiq gripper). 
}
\label{tab:inpaint_sim_result}
\vspace*{-5pt}
\end{table*}

\begin{table*}
\centering
{\color{black}
\begin{tabular}{c|c| ccc| ccc}
    \toprule
\multirow{2}{*}{\textbf{Task}}  & \multirow{2}{*}{\textbf{Source (Franka)}} & \multicolumn{3}{c}{\textbf{UR5e (Franka / Default Gripper)}}  & \multicolumn{3}{c}{\textbf{Kinova Gen3 (Franka / Default Gripper)}}  \\
    \cmidrule(lr){3-5} \cmidrule(lr){6-8}  
        & & Oracle  &  \algoname\,-- CP & \algoname  & Oracle  & \algoname\,-- CP & \algoname         \\
    \midrule
    \midrule
     Lift & 99\% & 99\% / 96\% & 85\% / 27\% & 96\% / 96\%  &  100\% / 99\% & 74\% / 14\% & 88\% / 96\% \\
    
    Stack & 97\% & 92\% / 93\% & 0\% / 0\%  & 80\% / 48\% & 95\% / 96\% & 7\% / 0\% & 76\% / 52\% \\
    
    Can & 60\% & 66\% / 51\% & 0\% / 0\% & 32\% / 24\%  & 51\% / 36\% & 2\% / 0\% & 24\% / 20\% \\
    
    Two Piece Assembly & 90\% & 88\% / 87\% &  33\% / 17\% & 64\% / 40\% &  95\% / 85\% & 38\% / 7\% & 76\% / 60\% \\
    
    Square & 77\% & 70\% / 48\% & 44\% / 3\% & 40\% / 20\% & 63\% / 27\% & 19\% / 0\% &  44\% / 20\% \\
    \bottomrule 
\end{tabular}
}
\caption{\NEW{\textbf{\algoname Results on Transferring Vision-Only Policies in Simulation.} For each task and robot arm combination, the source BC-RNN policy uses only the images as inputs. ``Oracle'' assumes access to a ground truth rendering of the source robot given the state of the target robot. \algoname\,-- CP ablates cross-painting and directly passes the visual observation of the target robot to the policy. \algoname uses cross-painting to generate the visual inputs for the policy. For each method, the first number represents the success rate when the target robot uses the source robot (Franka) gripper and the second number corresponds to using the target robot's default gripper (Robotiq gripper).}
}
\label{tab:inpaint_sim_result_no_proprio}
\vspace*{-5pt}
\end{table*}

We first study the effect of cross-painting on mitigating the visual gap between robots. We focus on closed-loop policies. Based on Table~\ref{tab:control-study-table}, we choose UR5e and Kinova Gen3 as two representative robots that show high potential for reusing the source robot policies. \NEW{The end-effector poses $p_{\mathrm{r}}^\mathcal{S}$ and $p_{\mathrm{r}}^\mathcal{T}$ are aligned in Robosuite by default. To study the importance of cross-painting, we apply \algoname to policies that take in both states and images as inputs, as well as policies that only take in images as inputs.} We use the ground-truth forward dynamics and \NEW{compare the success rates of the target robot using cross-painting (``\algoname''), with no cross-painting (``\algoname\,-- CP''), and with ground-truth oracle rendering (``Oracle''). }

To create a sim-to-sim gap, we choose Gazebo~\cite{gazebo} as our renderer. We manually position spotlights in the Gazebo environment to simulate the lighting in Robosuite for similar robot renderings. Similar to Sec.~\ref{sec:sim_study}, we first train the source robot policies with behavior cloning (BC-RNN) on the provided demonstration data for each task (200 demos for Lift and Can from Robomimic~\cite{robomimic2021}, 1000 demos for the other tasks from MimicGen~\cite{mandlekar2023mimicgen}), using the LSTM architecture with the ResNet-18 visual encoder~\cite{robomimic2021}. The policies utilize 84x84 images, and \algoname operates at approximately 40 Hz to cross-paint the images. 

\NEW{Table~\ref{tab:inpaint_sim_result} shows the results for policies that take in both states and images as observations, and Table~\ref{tab:inpaint_sim_result_no_proprio} shows those for policies that take in images only.} Similar to Table~\ref{tab:control-study-table}, transferring between robots but with the same gripper has higher success rates than transferring to a different gripper in most cases. 
For both UR5e and Kinova Gen3, \algoname has a performance drop of less than 25\% compared to querying the source policy on an oracle rendering for most tasks. 
In all cases, \algoname significantly outperforms the naive 0-shot performance without any visual gap mitigation. This suggests that cross-painting can effectively bridge the visual differences of the robots. 

\begin{figure*}
    \centering
    \includegraphics[width=\linewidth]{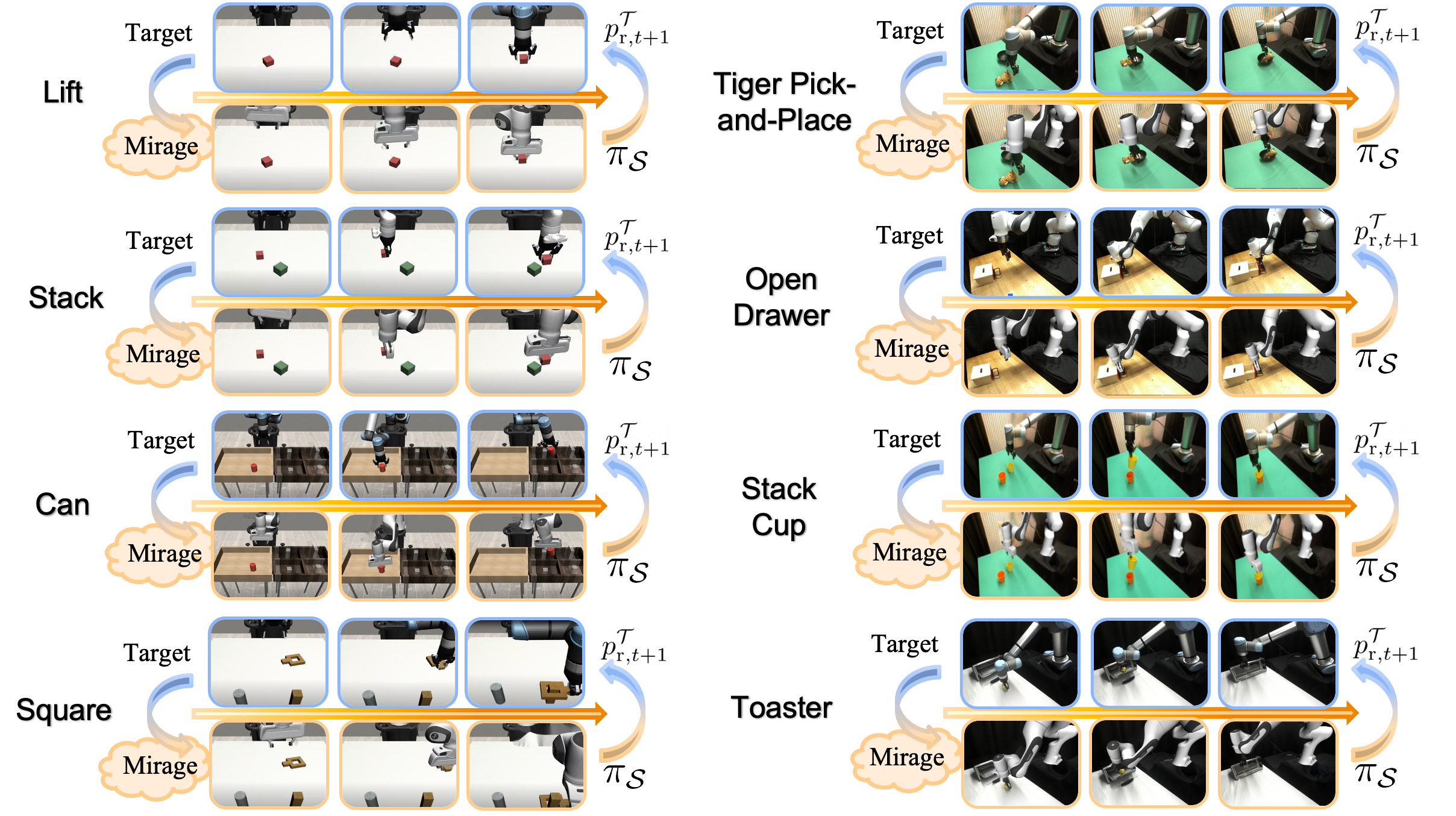}
    \vspace{-5pt}
    \caption{\NEW{\textbf{Trajectory Rollouts of Simulated (Left) and Real (Right) Tasks.} For each task, the top row shows the actual observations of the target robot during the trajectory rollout, and the bottom row shows the cross-painted images generated by \algoname that are passed to the source robot policy $\pi_\mathcal{S}$ to obtain the desired target robot pose $p_{r, t+1}^\mathcal{T}$.}}
    \label{fig:real_tasks}
\end{figure*}

\subsection{Physical Experiments}\label{subsec:physical_exps}

\begin{table*}
\centering
{\color{black}
\begin{tabular}{c|ccc|ccc|ccc|ccc}
    \toprule
\multirow{2}{*}{\diagbox[width=10em,trim=l]{\textbf{Policies}}{\textbf{Tasks}}} & \multicolumn{3}{c}{\textbf{Tiger in Bowl (Robotiq)}} & \multicolumn{3}{c}{\textbf{Open Drawer (Franka)}}  & \multicolumn{3}{c}{\textbf{Stack Cup (Franka)}} & \multicolumn{3}{c}{\textbf{Toaster (Robotiq)}}  \\
    \cmidrule(lr){2-4} \cmidrule(lr){5-7} \cmidrule(lr){8-10}  \cmidrule(lr){11-13}  
     & Source & T Grip & T Rob & Source & T Grip & T Rob & Source & T Grip & T Rob & Source & T Grip & T Rob        \\
    \midrule
    \midrule
    \multicolumn{13}{l}{\textbf{Policies that use both states and third-person images as observations:}} \\
    
    $\pi_\mathcal{S}$ (Diffusion Policy) &  90\% &  20\% & 0\% & 80\% & 0\%  &   0\% & 80\% & 0\%  &   0\% & 60\% & 0\% & 0\% \\
    
    $\pi_\mathcal{S}$\medspace + Cross-Painting (CP) & - & 50\% & 0\% & -  & 10\%  &  0\%  & -  &  20\% & 0\%  &  -  & 0\%  &  0\%  \\
    
    $\pi_\mathcal{S}$\medspace + State Alignment (SA) & - & 60\% & 50\% & -  & 0\%  & 0\% & - & 10\%  & 0\%  & - & 20\% & 0\% \\
    $\pi_\mathcal{S}$\medspace + SA + CP (\algoname)  & - & \textbf{90\%}   &  \textbf{90\%}  &   -  &  \textbf{70\%}  &  \textbf{60\%}   & - & \textbf{60\%}  &  \textbf{50\%}  &  -   & \textbf{40\%} & \textbf{30\%}  \\
    \midrule
    Octo  &  70\% &  0\% & 0\% & 40\% & 0\%  &   0\% & 30\% & 0\% & 0\% &  20\% & 0\% & 0\%  \\
    Octo + State Alignment (SA) &  - &  70\% & 10\% & - & 20\%  &   0\% & - & 20\% & 0\% &  - & 0\% & 0\%  \\
    \toprule
    \multicolumn{13}{l}{\textbf{Policies that use only third-person images as observations:}} \\
    $\pi_\mathcal{S}$ (Diffusion Policy) & 90\% & 20\% & 0\% & 70\%  & 0\%  & 0\% & 70\% & 0\%  & 0\%  & 50\% & 0\% & 0\% \\
    $\pi_\mathcal{S}$\medspace + CP (\algoname)  & - & \textbf{90\%}  & \textbf{60\%} & - &  \textbf{60\%}  &  \textbf{60\%}   & - & \textbf{60\%}  &  \textbf{40\%}  &  -   & \textbf{30\%} & \textbf{20\%}  \\
    \midrule
    \bottomrule 
\end{tabular}
}
\caption{{\color{black}\textbf{\label{real-inpaint-table} \algoname Results on Transferring Vision-Based Policies in Real with a Third-Person Camera.} We evaluate \algoname on 4 tasks in 2 settings: \textbf{Target (Different) Gripper (``T Grip''):} Transferring policies between the Franka gripper and the Robotiq 2F-85 gripper on a Franka robot. \textbf{Target (Different) Robot (``T Rob''):} Using the Franka with either gripper as the source robot (indicated in the parentheses in the first row), and the UR5 robot with the Robotiq gripper as the target robot. \textbf{Baseline $\pi_\mathcal{S}$ (Diffusion Policy):} Separate Diffusion Policy models~\cite{chi2023diffusionpolicy} trained on the source robot data for each task and evaluated on the source robot or zero-shot on the target embodiments. \textbf{Octo:} Octo Base model~\cite{octo_2023} finetuned on the source robot data from all tasks together and evaluated on the source robot or zero-shot on the target embodiments. \textbf{\algoname}: Evaluation of zero-shot transfer to the target embodiments using \algoname with the source policy being the corresponding baseline Diffusion Policy models. For policies that take in both states and images as inputs, \algoname applies both state alignment (SA) and cross-painting (CP). For policies that take in only images as inputs, \algoname applies cross-painting only. We perform ablations to study the importance of state alignment and cross-painting.
}
}
\label{tab:inpaint_real_result_with_octo}
\vspace*{-5pt}
\end{table*}

\begin{table}
\centering
{\color{black}
\begin{tabular}{c|cc|cc}
    \toprule
\multirow{2}{*}{\diagbox[width=10em,trim=l]{\textbf{Policies}}{\textbf{Tasks}}} & \multicolumn{2}{c}{\textbf{Tiger (Robotiq)}} & \multicolumn{2}{c}{\textbf{Drawer (Franka)}} \\
    \cmidrule(lr){2-3} \cmidrule(lr){4-5}  
     & Source & T Grip & Source & T Grip     \\
    \midrule
    \midrule
    \multicolumn{5}{l}{\textbf{Wrist Camera Only:}} \\
    
    $\pi_\mathcal{S}$ (Diffusion Policy) &  90\% &  0\% & 70\% & 0\%   \\
    
    $\pi_\mathcal{S}$\medspace + Cross-Painting (CP) & - & 10\% & -  & 0\%    \\
    
    $\pi_\mathcal{S}$\medspace + State Alignment (SA) & - & 0\%  & -  & 0\%  \\
    $\pi_\mathcal{S}$\medspace + SA + CP (\algoname)  & - & \textbf{70\%}   &   -  &  \textbf{40\%}     \\
    \midrule
    \multicolumn{5}{l}{\textbf{Wrist Camera + Third-Person Camera:}} \\
    
    $\pi_\mathcal{S}$ (Diffusion Policy) &  90\% &  0\% & 90\% & 0\%   \\
    
    $\pi_\mathcal{S}$\medspace + Cross-Painting (CP) & - & 0\% & -  & 0\%   \\
    
    $\pi_\mathcal{S}$\medspace + State Alignment (SA) & - & 0\%  & -  & 0\%  \\
    $\pi_\mathcal{S}$\medspace + SA + CP (\algoname)  & - & \textbf{80\%}   &   -  &  \textbf{60\%}     \\
    \midrule
    \bottomrule 
\end{tabular}
}
\caption{{\color{black}\textbf{\label{real-inpaint-table} \algoname Results on Transferring Policies in Real that Use a First-Person (Wrist) Camera.} We evaluate \algoname for transferring between grippers on 2 tasks, with the source gripper indicated in the parentheses in the first row. \textbf{Target (Different) Gripper (``T Grip''):} Transferring policies between the Franka gripper and the Robotiq 2F-85 gripper on a Franka robot. We study policies that use only the wrist camera or both wrist and third-person cameras.  We perform ablations to study the importance of state alignment (SA) and cross-painting (CP).
}
}
\label{tab:inpaint_real_result_wrist_cam}
\vspace*{-5pt}
\end{table}

We evaluate \algoname across 3 different embodiments, Franka with Franka and Robotiq 2F-85 grippers, and UR5 with Robotiq 2F-85 gripper. We evaluate on 4 manipulation tasks: (1) Pick up a stuffed animal (tiger) and put it into a bowl~\cite{BerkeleyUR5Website}, (2) open a toy drawer~\cite{walke2023bridgedata}, (3) stack one cup into another~\cite{BerkeleyUR5Website}, and (4) put a pepper into a toaster and close its glass door~\cite{octo_2023}. We select Franka as the source robot. For tasks (2) and (3), the source robot uses the Franka gripper, while for the others, the source robot is equipped with the Robotiq gripper.

We study policy transfer in 2 different settings. The first setting is transferring between grippers only on the Franka robot. The second setting involves robot transfer, where we use the Franka with either gripper as the source robot, and the UR5 with the Robotiq 2F-85 as the target robot. \NEW{We use a ZED 2 camera positioned from the side for each robot, and a ZED Mini as the wrist camera for each gripper.}

For each task, we first use an Oculus controller to collect between 200-400 human demonstration trajectories on the source robot using VR teleoperation at 15 Hz~\cite{khazatsky2024droid}. \NEW{We then train 2 separate diffusion policies~\cite{chi2023diffusionpolicy} for each task, one that uses both states and images as the inputs, and one that uses only images as inputs. They serve as the source policies that we seek to transfer.}  On the target UR5 robot, we place the camera(s) at similar poses, with up to 4 cm differences from those in the source setup. Similar to Sec.~\ref{subsec:sim_experiments}, we use Gazebo as our renderer. We use a per-dimension linear forward dynamics model and use the demonstration data to fit the regression coefficients. We use a blocking controller on the UR5 to reach the desired poses.

\NEW{We note that on the Franka robot, there is a 45$^\circ$ difference in rotation between the Franka gripper and Robotiq gripper in their installations. Additionally, there is a 90$^\circ$ rotational difference in the coordinate frames of the Robotiq gripper in the base frame of the UR5 robot compared to that of the Franka robot. To study the value of state alignment (SA) $p_{\mathrm{r},t+1}^\mathcal{T} = T_{\mathcal{S}}^{\mathcal{T}} p_{\mathrm{r},t+1}^\mathcal{S}$ and cross-painting (CP) $o^{\mathcal{T}\rightarrow\mathcal{S}}$, we conduct ablations of both modules from \algoname. Additionally, as a baseline, we also directly apply the source policy $\pi_\mathcal{S}$ (the task-specific diffusion policy trained on the source robot only) to the target embodiment for each task, without accounting for either coordinate frame or robot visual differences.}

\begin{figure}[t]
\center
\includegraphics[width=0.5\textwidth]{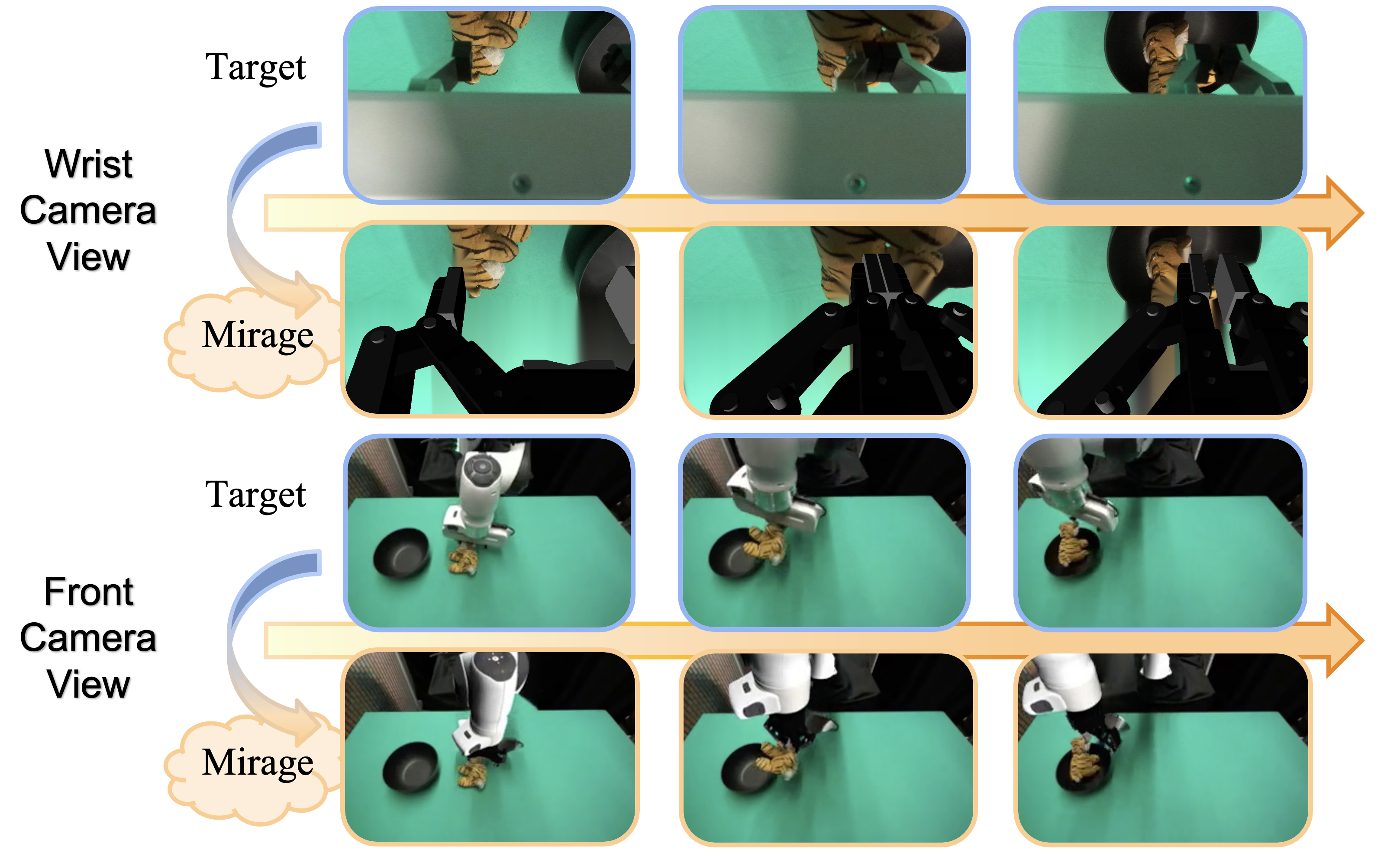}
\caption{
\NEW{\algoname applied to first-person wrist camera images and third-person front camera images. For each view, the top row shows the actual observations of the target robot during the rollout, and the bottom row shows the cross-painted images generated by \algoname.}
}
\vspace{-5pt}
\label{fig:front_wrist_view}
\end{figure}

\NEW{We also compare \algoname to a finetuned multi-task policy, Octo~\cite{octo_2023}. 
Octo is a recently released state-of-the-art generalist policy for robotic manipulation, pre-trained on 800k robot episodes from the Open X-Embodiment dataset~\cite{open_x_embodiment_rt_x_2023}, including trajectories collected on Franka and UR5. We take the pretrained Base model (93M parameters) and finetune all weights on the combined 4-task demonstration data of the source robot with language goal conditioning. It uses a third-person camera image, and the proprioception input is the Cartesian position of the end effector and the gripper position. 
}

Table~\ref{tab:inpaint_real_result_with_octo} shows the results for each task on all 3 robot setups: Source (Franka robot + Franka/Robotiq gripper), Different Gripper (Franka robot + the other gripper), and Different Robot (UR5 + Robotiq gripper). We can see that, for both gripper transfer and robot (and gripper) transfer, \algoname achieves strong zero-shot performance, significantly outperforming all baselines. \NEW{Specifically, for policies that use both states and third-person images as inputs, state alignment and cross-painting are both crucial for task success on the target embodiments. This is not surprising since both transformations are needed to bring the test-time input back to the distribution of the training data. Comparing the two base models with state alignment only, task-specific diffusion policies achieve stronger performance on the source robot it is trained on, while Octo achieves better success rates on most tasks when the grippers are swapped, suggesting slightly more robustness to the visual changes. Still, Octo tends to miss more grasps or gets stuck, resulting in a lower success rate than \algoname by 20\% to 50\%. When being evaluated on the UR5, no baselines are able to achieve any success on the 3 more difficult tasks. For policies that take in only images as inputs, state alignment is unnecessary, and cross-painting alone can achieve strong results on the target embodiment, with only minor performance degradation compared to the source robot.}

Additionally, similar to Tables~\ref{tab:control-study-table} and \ref{tab:inpaint_sim_result}, the stronger the source policy is, the smaller the performance gap is during transfer. Interestingly, unlike Tables~\ref{tab:control-study-table} and \ref{tab:inpaint_sim_result}, we do not observe that transferring between robots with the same gripper achieves higher performance than transferring between grippers only. We hypothesize that this is due to the minor scene setup differences that cause a natural drop in policy performance. On the other hand, the failure modes we observe on the different robots or grippers are all very similar to those from the source policy on the source robot, such as not dropping the cup high enough during cup stacking or not keeping the gripper low enough when closing the door of the toaster oven.

\NEW{We also evaluate \algoname on first-person images. We select the ``Tiger'' and ``Drawer'' tasks and train a diffusion policy with wrist camera inputs and a diffusion policy that take both first-person and third-person camera inputs. Both models use proprioceptive inputs. For the latter, \algoname applies cross-painting to both images. From Table~\ref{tab:inpaint_real_result_wrist_cam}, we can see that only \algoname is able to achieve high success rates on the target embodiments. We hypothesize that this is because there is a large visual difference between the 2 grippers (see Figure~\ref{fig:front_wrist_view}), which confuses the policy if cross-painting is not applied to bridge the gap. This shows that \algoname is flexible with the number of cameras as well as the camera view.}

\subsection{Sensitivity Analysis of \algoname}\label{subsec:sensitivity}

\begin{table}
\centering
\begin{tblr}{
  row{2} = {c},
  row{3} = {c},
  row{4} = {c},
  row{5} = {c},
  row{6} = {c},
  row{7} = {c},
  row{8} = {c},
  row{9} = {c},
  row{10} = {c},
  row{11} = {c},
  cell{1}{1} = {c},
  cell{1}{2} = {c},
  cell{1}{3} = {c},
  cell{2}{1} = {r=4}{},
  cell{2}{2} = {r=2}{},
  cell{4}{2} = {r=2}{},
  cell{6}{1} = {r=4}{},
  cell{6}{2} = {r=2}{},
  cell{8}{2} = {r=2}{},
  vline{2-4} = {1-13}{},
  vline{4} = {3-12}{},
  vline{3-4} = {4,6,8,10,13}{},
  hline{1-2,6,10} = {-}{},
  hline{4,8} = {2-4}{},
}
\textbf{Component}     & \textbf{Factor}               & \textbf{Variation}                  & \textbf{Success Rate} \\
\textbf{Visual}        & Calibration Error  & \textasciitilde{}10 pixels & 80\%         \\
              &                    & \textasciitilde{}30 pixels & 50\%         \\
              & Luminance          & +50                        & 90\%         \\
              &                    & -50                        & 80\%         \\
\textbf{Control}       & Gain               & x0.5                         & 90\%         \\
              &                    & x2                       & 60\%         \\
              & Offset             & 2\,cm                        & 70\%         \\
              &              & 5\,cm                        & 20\%   
\end{tblr}
\caption{\textbf{Sensitivity of \algoname Components on Policy Performance on the Tiger Pick-and-Place Task.} For the visual components, we study the effect of camera calibration error and the luminance of inpainted robot rendering. For the control components, we evaluate sensitivity of \algoname to the control gain of the target robot controller and the $z$-offset in the proprioceptive values.
}
\label{tab:sensitivity_analysis}
\vspace*{-10pt}
\end{table}

\begin{figure}[t]
\center
\includegraphics[width=0.5\textwidth]{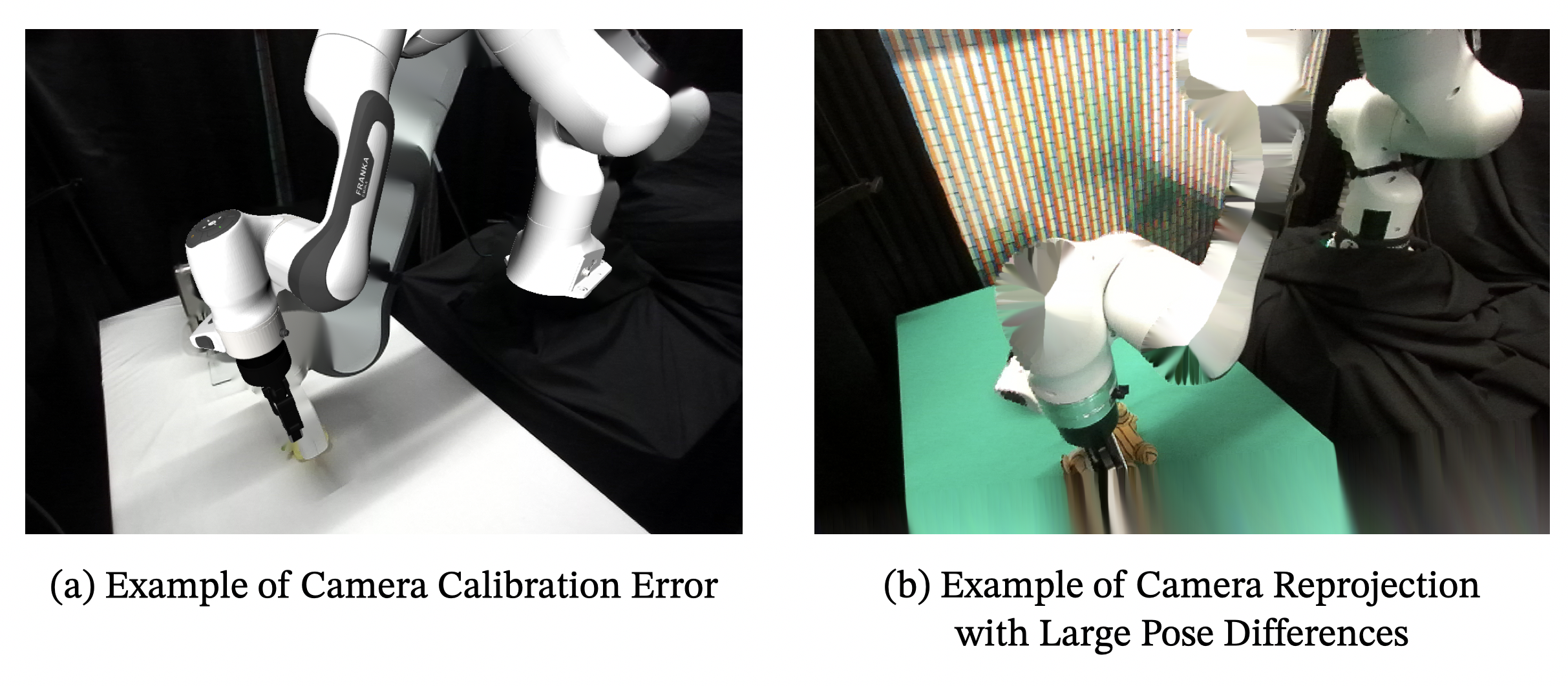}
\caption{
(a) An example of camera calibration error resulting in failure to mask all of the target robot out; (b) An example of the artifacts introduced due to large changes in camera angles.
}
\vspace{-5pt}
\label{fig:calibration_error_reproject_example}
\end{figure}

We examine the sensitivity of policy performance to the various components of \algoname using the Tiger-in-Bowl task.
For the visual components, we study the effect of camera calibration error on the target robot, \NEW{the quality of the robot rendering}, and background changes between the target and source robots. 
To study the effect of camera calibration \NEW{and URDF inaccuracy}, we add offsets to the masks and inpainted robots. As the calibration error increases, the segmentation masks become less accurate and leave parts of the robot unmasked. This leads to cross-painted images that include remnants of the target robot (see Fig.~\ref{fig:calibration_error_reproject_example}(a)). 
Table~\ref{tab:sensitivity_analysis} shows that large artifacts result in non-negligible policy performance degradation. For the sensitivity of the renderer quality, we adjust the inpainted pixels' luminance by an offset amount, and Table~\ref{tab:sensitivity_analysis} shows that the trained policy is relatively robust to it. 

We also examine the sensitivity to the control pipelines of \algoname. To simulate noises in the forward dynamics model and inaccuracy in the target robot's controller, we adjust the control gain and find that large values affect performance. Additionally, when the target robot's proprioceptive values are shifted due to offsets, the trained policy's performance drastically decreases. This is not surprising as the policy is trained with matching proprioceptive values and image observations, and large offsets in the proprioceptive values correspond to distribution shifts in the state input, which \algoname does not mitigate.

\NEW{Additionally, we evaluate the robustness of \algoname to changes in camera angles between the source and target robot setups, and in other words, the effectiveness of \algoname's camera reprojection step from the target image back to the source camera pose. 
We select 3 levels of camera pose differences:
\begin{itemize}
    \item Small Differences: Up to 1 cm in total translation and 5$^{\circ}$ in total rotation mismatches;
    \item Medium Differences: Up to 5 cm in total translation and 15$^{\circ}$ in total rotation mismatches;
    \item Large Differences: Up to 10 cm in total translation and 15$^{\circ}$ in total rotation mismatches;
\end{itemize}

\begin{table}[t]
\centering
\begin{tabular}{c| cc| cc}
    \toprule
\multirow{2}{*}{\textbf{Camera Pose Change}}  & \multicolumn{2}{c}{\textbf{Source}}  & \multicolumn{2}{c}{\textbf{Different Gripper}}  \\
    \cmidrule(lr){2-3} \cmidrule(lr){4-5}  
        & \scriptsize{w/o reproj.} & \scriptsize{\algoname}  & \scriptsize{w/o reproj.} & \scriptsize{\algoname}         \\
    \midrule
    \midrule
     Small (1 cm + 5$^\circ$) &   90\% &  90\%  &   90\% & 90\%  \\
    Medium (5 cm + 15$^\circ$) &  60\% &   80\%  &   50\%  &  80\%  \\
    Large (10 cm + 15$^\circ$) &  30\% &   50\%  &   20\% & 40\% \\
    \bottomrule 
\end{tabular}
\caption{\NEW{\textbf{\algoname Effectiveness of Camera Reprojection.} We evaluate \algoname with and without the camera reprojection on the Tiger Pick-and-Place task with 3 levels of camera pose changes: $\leq$ 1 cm and $\leq$ 5$^\circ$, $\leq$ 5 cm and $\leq$ 15$^\circ$, $\leq$ 10 cm and $\leq$ 15$^\circ$. Results suggest that, when there are medium to large differences between camera poses, applying camera reprojection is beneficial.}
}
\label{tab:camera_angle}
\vspace*{-5pt}
\end{table}

For each level of target camera angles, we compare the performance of the policy with and without the camera reprojection step (\algoname and ``w/o reproj.'' respectively) both on the source robot setup and on the Different Gripper setup. Specifically, for the source robot setup, no cross-painting is needed, so \algoname is effectively just camera reprojection. For the different gripper setup, cross-painting is applied after reprojection, while ``w/o reproj.'' directly applies cross-painting in the target camera pose instead of in the source camera pose. 

Table~\ref{tab:camera_angle} shows that applying camera reprojection has a 20-30\% improvement in the success rates when there are medium to large differences between the camera poses of the source and the target. When the difference increases, there are fewer points projected back to the source image frame, resulting in more missing pixels being inpainted. This leads to blurriness and artifacts close to the boundary of the images (see Fig.~\ref{fig:calibration_error_reproject_example}(b)), making it more likely for the policy to miss the grasp.}

\section{Conclusion} 
\label{sec:conclusion}
We propose \algoname, a novel algorithm for zero-shot transfer of manipulation policies to unseen robot embodiments. \algoname relies on two key ingredients: (1) using robot URDFs and renderers to mitigate the visual differences through cross-painting, and (2) choosing the end effector Cartesian pose as the state and action spaces and using knowledge of the robot coordinate frames to align the source and target robots. Through both simulation and physical experiments across 9 tasks, we find that \algoname can successfully enable zero-shot transfer across grippers and robot arms, significantly outperforming a state-of-the-art generalist policy.

\textbf{Limitations and future work.} 
To enable zero-shot transfer, \algoname relies on knowledge of the robots and setups such as the robot URDFs, coordinate frames, and camera matrices. Also, since we use a blocking OSC controller during execution, \algoname is most suitable for quasistatic tasks. 
Additionally, we do not consider transferring between different backgrounds. Future work can explore combining \algoname with orthogonal approaches such as augmentation and scaling to enable policies to be robust to background changes.

\section*{Acknowledgments}
This research was performed at the AUTOLab at UC Berkeley in affiliation with the Berkeley AI Research (BAIR) Lab, and the CITRIS ``People and Robots'' (CPAR) Initiative, and in collaboration with Google DeepMind. The authors are supported in part by donations from Google, Toyota Research Institute, and equipment grants from NVIDIA.  L.Y. Chen is supported by the National Science Foundation (NSF) Graduate Research Fellowship Program under Grant No. 2146752. We thank Zehan Ma, Simeon Adebola, Max Fu, Ethan Qiu, and Roy Lin for helping with physical experiments, Karl Pertsch and Sudeep Dasari for support with Octo model fine-tuning, Ajay Mandlekar for support with Robosuite, and Will Panitch and Qianqian Wang for valuable feedback.

\bibliographystyle{plainnat}
\bibliography{references}

\clearpage
\begin{appendices}
\section*{Appendix}
We provide the following additional results and details in this appendix:
\begin{itemize}
    \item Section~\ref{appendix:delta_control}: State-based transfer experiment results, using the same delta actions on the target robot instead of a blocking controller, and trajectory playback results;
    \item Section~\ref{appendix:cv2_inpaint_effects}: Additional implementation details of the cross-painting step of \algoname;
\end{itemize}

\begin{table*}[!ht]
\centering
{\begin{tabular}{c|c|ccccc|ccccc}
\toprule[1pt]
\multirow{2}{*}{\textbf{Task}} & \multirow{2}{*}{\textbf{Source (Franka)}} & \multicolumn{5}{c}{\textbf{Franka Gripper on Target Robot}}  & \multicolumn{5}{c}{\textbf{Default Gripper on Target Robot}} \\

    \cmidrule(lr){3-7} \cmidrule(lr){8-12}

& & UR5e & IIWA & Kinova Gen3 & Sawyer & Jaco & UR5e & IIWA & Kinova Gen3 & Sawyer & Jaco \\
\midrule[0.1pt]
\midrule[0.1pt]
Lift (Robosuite) & 100\% & 91\% & 94\% & 96\% & 98\% & 96\% & 99\% & 53\% & 98\% & 72\% & 63\% \\

Stack & 95\% & 36\% & 39\% & 74\% & 60\% & 36\% & 41\% & 28\% & 43\% & 40\% & 38\% \\

Can & 98\% & 87\% & 93\% & 89\% & 78\% & 86\% & 90\% & 25\% & 73\% & 41\% & 60\% \\

Two Piece Assembly & 96\% & 44\% & 65\% & 54\% & 43\% & 41\% & 44\% & 45\% & 48\% & 29\% & 15\% \\

Square & 81\% & 10\% & 15\% & 3\% & 5\% & 9\% & 6\% & 1\% & 2\% & 5\% & 0\% \\

\bottomrule[1pt]
\end{tabular}} 
\caption{\textbf{State-Based Transfer Experiment with Delta Actions.} We evaluate state-based policies trained on a Franka robot for each task across five different robots equipped either with the original Franka gripper or with each target robot's default gripper. The target robot executes the same delta Cartesian action as the source Franka robot instead of reaching the same absolute pose.}
\label{tab:control-study-table-delta} 
\end{table*}

\begin{table*}[htbp!]
\centering
{\begin{tabular}{c|cccccc}
\toprule[1pt]
\textbf{Task} & \textbf{Source (Franka)} & \textbf{UR5e} & \textbf{IIWA} & \textbf{Kinova Gen3} & \textbf{Sawyer} & \textbf{Jaco} \\
\midrule[0.1pt]
Lift (Robosuite) & 100\% & 100\% & 99\% & 100\% & 85\% & 90\% \\

Stack & 100\% & 100\% & 100\% & 89\% & 76\% & 50\% \\

Can & 100\% & 80\% & 75\% & 50\% & 80\% & 54\% \\

Two Piece Assembly & 100\% & 98\% & 90\% & 94\% & 93\% & 69\% \\

Square & 100\% & 73\% & 43\% & 27\% & 82\% & 2\% \\

\bottomrule[1pt]
\end{tabular}}
\caption{\textbf{Success Rates of Target Robots Playing Back the Source Robot Trajectories.} Instead of executing the source robot policy, we evaluate the success rates of target robots performing trajectory playback. Specifically, we use the robot's blocking controller with the achieved pose of the source robot as the target pose instead of performing the same delta actions.}
\label{tab:control-study-trajectory-playback}
\end{table*}

\section{Delta Control and Trajectory Playback}\label{appendix:delta_control}
In Section~\ref{subsec:sim_study_results} Table~\ref{tab:control-study-table}, we see that target robots can achieve high task success rates by querying the source robot policy and referencing the source robot for desired poses. Specifically, we use a blocking controller with absolute pose commands. In this section, we compare and experiment with executing the same delta action from the policy directly on the target robot with a non-blocking controller. Table~\ref{tab:control-study-table-delta} shows that there is a significant drop in performance, indicating that the difference in the forward dynamics between robots cannot be ignored when transferring policies and that leveraging a blocking controller on the target robot is an effective way to mitigate this difference.

Additionally, we compare querying a source robot policy with playing back successful source robot trajectories. Since we have shown that executing the same delta actions as the source robot does not transfer well, we use the achieved poses by the source robots in the demonstrated trajectories paired with blocking Cartesian controllers on the target robots. Table~\ref{tab:control-study-trajectory-playback} shows the results. We see that, compared with Table~\ref{tab:control-study-table}, the performances are similar to those using the source robot policies. This suggests that these tasks can be achieved by those target robots using the same strategy. In fact, this is similar to the data generation strategy adopted by MimicGen~\cite{mandlekar2023mimicgen}, in which the authors roll out the (transformed) trajectories on the target robots and filter out those unsuccessful ones and then train a model on the rest. Comparing playing back human demonstration trajectories and querying policies on the source robot, we observe that in some cases where the learned policies are brittle or suboptimal, they leave less room for error and are less robust to changes in the gripper or tracking errors. On the other hand, good learned policies can potentially adapt to tracking errors, adjust their poses, and retry grasps if necessary, thus outperforming trajectory playback in some cases.

\section{Additional Implementation Details of \algoname}\label{appendix:cv2_inpaint_effects}

In this section, we detail the inpainting step in the cross-painting process of \algoname, and its sensitivity to calibration errors. Once we have used the segmentation mask provided by Gazebo to black out the target robot, we use Fast Marching \cite{telea2004image} algorithm to fill in the black pixels. Specifically, we use a neighborhood radius of 3 pixels and the Telea inpainting method, which is based on a fast marching algorithm \cite{telea2004image}. While the filled-in pixels appear blurry, they generally match the color scheme of the rest of the image when the calibration error is low. However, with higher calibration errors, the segmentation masks become less accurate, resulting in remnants of the target robot in the cross-painted images (see Fig.~\ref{fig:calibration_error_reproject_example}(a)). To mitigate this issue, we dilate the segmentation masks to ensure that the target robot is completely blacked out even with calibration errors. With moderate to low calibration errors, we dilate the robot arm segmentation mask with a 3x3 kernel for 20 iterations and the gripper segmentation mask for 10 iterations. However, when the calibration error is higher, we dilate the arm and gripper segmentation masks by 40 and 20 iterations respectively. The gripper mask is dilated significantly less than the robot arm mask because too much dilation would blur the objects that the robot needs to interact with, leading to worse performance.

\end{appendices}

\end{document}